%% file: main_v3.tex
\documentclass{article}
\usepackage{spconf,amsmath,graphicx,multicol}
\usepackage{algorithm2e}

\usepackage{amsfonts}
\usepackage{amssymb}
\usepackage{tabularx}

\usepackage{capt-of}
\usepackage{bm}

\newenvironment{Table}
  {\par\bigskip\noindent\minipage{\columnwidth}\centering}
  {\endminipage\par\bigskip}
\newcommand{\mbf}[1]{\ensuremath{\mathbf{#1}}}

\newcommand*\tageq{\refstepcounter{equation}\tag{\theequation}}

\title{Triplet Similarity Embedding for Face Verification}
\name{Swami Sankaranarayanan$^{1,2}$, Azadeh Alavi$^2$, Rama Chellappa$^{1,2}$}

\address{ $^1$Department of Electrical and Computer Engineering, \\
$^2$Center for Automation Research, UMIACS, University of Maryland, College Park, MD 20742\\
{\small{\{swamiviv, azadeh, rama\}@umiacs.umd.edu}}
\thanks{This research is based upon work supported by the Office of the Director of National Intelligence (ODNI), Intelligence Advanced Research Projects Activity (IARPA),via IARPA R\&D Contract No. 2014-14071600012.  The views and conclusions contained herein are those of the authors and should not be interpreted as necessarily representing the official policies or endorsements, either expressed or implied, of the ODNI, IARPA, or the U.S. Government.  The U.S. Government is authorized to reproduce and distribute reprints for Governmental purposes notwithstanding any copyright annotation thereon.}
}
\begin{document}
\RestyleAlgo{boxruled}
\LinesNumbered

%
\parskip 0pt
\maketitle
\begin{abstract}
In this work, we present an unconstrained face verification algorithm and evaluate it on the recently released IJB-A dataset \cite{ijba15} that aims to push the boundaries of face verification methods. The proposed algorithm couples a deep CNN-based approach with a low-dimensional discriminative embedding learnt using triplet similarity constraints in a large margin fashion. Aside from yielding performance improvement, this embedding provides significant advantages in terms of memory and post-processing operations like hashing and visualization. Experiments on the IJB-A dataset show that the proposed algorithm outperforms state of the art methods in verification and identification metrics, while requiring less training time.
\end{abstract}

\input{intro2}
\input{background2}
\input{network2}

\input{tl3}
\input{results3}

\input{conclusion3}

\bibliographystyle{IEEEbib}
\bibliography{refs}

\end{document}

%% file: intro2.tex
\section{Introduction}
\label{sec:intro}
Recently, with the advent of curated face datasets like Labeled faces in the Wild (LFW) \cite{lfw} and advances in learning algorithms like Deep neural nets, there is more hope that unconstrained face verification problems can be solved.. A face verification algorithm compares two given templates that are typically not seen during training. Research in face verification has progressed very well over the past few years, resulting in the saturation of performance in the LFW dataset, yet the problem of unconstrained face verification remains a challenge. This is evident by the performance of traditional algorithms in the publicly available IJB-A dataset (\cite{ijba15}, \cite{fvff15}) that was released recently. In addition, despite the superb performance of CNN-based approaches compared to traditional methods, a drawback of such methods is the lengthy training time. In this work, we provide a Deep CNN (DCNN) architecture that ensures faster training, and investigate how much the performance can be improved if we are provided with domain specific data. Specifically our contributions are as follows:
\begin{itemize}
\item We propose a deep network architecture and a training scheme that ensures faster training time.
\item We formulate a triplet similarity embedding learning method to improve the performance of deep features.
\end{itemize}

\begin{figure}
\includegraphics[width=0.5\textwidth]{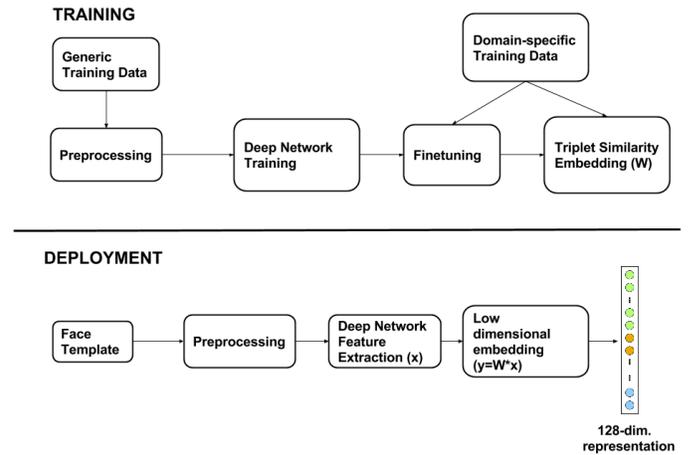}
\caption{Training and Deployment processing pipeline}
\label{fig:pipeline}
\end{figure}

Figure \ref{fig:pipeline} shows the processing pipeline used in this work. During training, we use a generic dataset (that is independent of the IJB-A dataset) to train our deep architecture. To further improve performance, we use the training data provided with the IJB-A data to finetune our DCNN model. We learn the proposed triplet similarity embedding using the same training set but with features extracted from the finetuned DCNN model. During the deployment phase, given a face template, we extract the deep features using the finetuned model after some preprocessing steps. The deep features are projected into a low-dimensional space using the embedding matrix learnt during training (note that the projection involves only matrix multiplication). We use the 128-dimensional feature as the final representation of the given face template. 

This paper is organized as follows: Section \ref{SOA} places our work among the recently proposed approaches for face verification. Section \ref{net} details the network architecture and training scheme. The triplet similarity embedding learning method is described in Section \ref{tse} followed by results on the IJB-A dataset and discussion in Section \ref{sec:results}.

%% file: background2.tex
\section{Related Work}\label{SOA}
This work broadly consists of two components: the deep network used as a feature extractor and the learning procedure that projects the input features into a discriminative low-dimensional space. In the past few years, there have been numerous works in using deep features for tasks related to face verification. DeepFace \cite{deepface14} uses a carefully crafted 3D alignment procedure to preprocess face images and feeds them to the a deep network (with 120M parameters) that is trained with a large training set. A kernel classifier is trained on the resulting features to make the final verification decision. More recently, Facenet \cite{facenet15} uses the inception architecure and a large private dataset to train a deep network using a triplet distance loss function. The training time for this network is of the order of few weeks. Since the release of the IJBA dataset \cite{ijba15}, there have been several works that have published results on the verification protocol. Most notably, \cite{iccv15} and \cite{wang15} train deep networks using the CASIA-WebFace dataset \cite{casia14} with training time of the order of few days. This work proposes a network architecture and a training scheme that needs about 20 hrs of training time. Similar to \cite{iccv15}, we train our deep network using the CASIA-Webface dataset.

The idea of learning a compact and discriminative representation has been around for decades. Weinberger \emph{et al.}  \cite{lmnn05} introduced the idea of using an Semi Definite Programming (SDP)-based formulation to learn a metric satisfying pairwise and triplet distance constraints in a large margin framework. More recently, this idea has been successfully applied to face verification by integrating the loss function within the deep network architecture (\cite{facenet15}, \cite{parkhi15}). Joint Bayesian metric learning has been another popular metric used for face verification (\cite{fvf13},\cite{chen15wacv}). These methods either require a large dataset for convergence or learn a metric directly thereby not enabling further operations like discriminative clustering or hashing. The current work formulates an optimization problem based on triplet similarities that converges well and similar to \cite{facenet15}, gives a compact representation of the input. The embedding scheme described in this work is a more general framework that can be applied to any general setting where labeled training data is available.

%% file: network2.tex
\section{Network Architecture}\label{net}

This section details the architecture and training algorithm for the deep network used in our work. The architecture of our network is shown in Figure \ref{arch}. Our architecture closely follows the architecture of the AlexNet \cite{alexnet12} with the following differences:

\begin{itemize}
\item The f/c layers have fewer parameters thus reducing the number of parameters by more than a half.
\item We use Parametric Rectifier Linear units (PReLU's) instead of ReLU, since they allow a negative value for the output based on a learnt threshold and have been shown to improve convergence rate \cite{prelu15}.
\end{itemize}

The main reason for mirroring the AlexNet architecure in the convolutional layers is closely tied to the training scheme: We initialize the convolutional layer weights with the weights from the AlexNet model that was trained on the ImageNet challenge dataset. Several recent works (\cite{transfer1},\cite{transfer2}) have empirically shown that this transfer of knowledge across different networks, albeit for a different objective, improves performance and more significantly reduces the need to train over a large number of iterations. To learn more domain specific information, we add an additional convolutional layer, \textit{conv6} and initialize the fully connected layers \textit{fc6-fc8} from scratch. Since the network is used as a feature extractor, the last layer \textit{fc8} is removed during deployment, thus reducing the number of parameters to 15M. When the network is deployed. the features are extracted from \textit{fc7} layers resulting in a dimensionality of 512.

\begin{Table}

\centering

\begin{tabular}{|c|c|c|}
 \hline
 \multicolumn{3}{|c|}{Deep Network Architecture} \\
 \hline
 Layer & Kernel Size/Stride & \#params\\
 \hline
 conv1    &11x11/4 & 35K\\
 conv2    &5x5/2 & 614K\\
 conv3    &3x3/2 & 885K\\
 conv4    &3x3/2 & 1.3M\\
 conv5    &3x3/1 & 885K\\
 conv6    &3x3/1 & 590K\\
 fc6    &1024 & 9.4M \\
 fc7    &512 & 524K \\
 fc8    &10575 & 5.5M \\
 Softmax Loss & &  Total: 19.8M\\

 \hline

\end{tabular}
\captionof{table}{Deep Network architecture details}
\label{arch}
\end{Table}

%% file: tl3.tex
\section{Learning a discriminative embedding}\label{tse}
In this section, we describe a low-dimensional embedding learnt using data such that the resulting projections are more discriminative. Aside from an improved performance, this embedding provides significant advantages in terms of memory, post-processing operations like hashing and visualization.

Consider a triplet $\{a,p,n\}$, where $a$ (anchor) and $p$ (positive) are from the same class, but $n$ (negative) belongs to a different class. Our objective is to learn a linear projection $\mbf{W}$ from the data such that the following constraint is satisfied:
\begin{align*}
(\mbf{W}a)^{T} \cdot (\mbf{W}p) > (\mbf{W}a)^{T} \cdot (\mbf{W}n)
\tageq \label{eq:constraint}
\end{align*}

In our case, $\{a,p,n\} \in \mbf{R}^{512}$ are deep descriptors which are normalized to unit length. As such, $(\mbf{W}a)^T \cdot (\mbf{W}p)$ is the dot-product or the similarity between ${a,p}$ under the projection $\mbf{W}$. The constraint in (\ref{eq:constraint}) requires that the similarity between the anchor and positive samples should be higher than the similarity between the anchor and negative samples in the low dimensional space represented by $\mbf{W}$. Thus, the mapping matrix $\mbf{W}$ pushes similar pairs closer and dissimilar pairs apart, with respect to the anchor point. By choosing the dimensionality of $\mbf{W}$ as $d \times 512$ where $d < 512$, we achieve dimensionality reduction in addition to better performance. For our work, we fix $d=128$ based on cross validation.

Given a set of labelled data points, we solve the following optimization problem: 
\begin{align*}
\underset{\mbf{W}}{\text{argmin}} \sum_{ {a,p,n} \in \mathbb{T}} max(0,\alpha + a^T\mbf{W^TW}n-a^T\mbf{W^TW}p) 
\tageq \label{eq:train1}
\end{align*}
where $\mathbb{T}$ is the set of triplets and $\alpha$ is a margin parameter chosen based on the validation set. In practice, the above problem is solved in a Large-Margin framework using Stochastic Gradient Descent (SGD) and the triplets are sampled online. The update step for solving (\ref{eq:train1}) with SGD is:

\begin{align*}
\mbf{W}_{t+1} = \mbf{W}_t - \eta * \mbf{W}_t * (a(n-p)^T  + (n-p)a^T)
\tageq \label{eq:update}
\end{align*}
where $\mbf{W}_t$ is the estimate at iteration $t$, $\mbf{W}_{t+1}$ is the updated estimate, $\{a,p,n\}$ is the triplet sampled at the current iteration and $\eta$ is the learning rate.

The pseudo-code for obtaining $\mbf{W}$ is shown in Algorithm \ref{algo:1}. $\mbf{W}$ is initialized with the first $d$ principal components of the training data. At each iteration, a random anchor and a random positive data points are chosen. To choose the negative, we perform hard negative mining, ie. we choose the data point that most violates the constraint in (\ref{eq:train1}) among the randomly chosen 2000 negative instances at each iteration. 

\begin{algorithm}
 \KwData{Data-labels pairs: $\{x_i,y_i\}_{i=1}^{N}$, Learning rate $\eta$}
 \KwResult{Embedding matrix, W $\in$ $\mbf{R}^{d_{in} \times d_{out}}$ }
 \tcc{initialization}
 $\mbf{W_0} = pca(\mbf{X},'dout')$\;
 $t$ $=$ 0 \;
 \While{t $<$ maxIter}{
  a = get-random-anchor\;
  p = get-random-positive\; 
  n = get-hard-negative\;
  \If{(a,p,n) violates (\ref{eq:train1}) }{
   Update $\mbf{W}$ according to (\ref{eq:update})\;
   }
   $t$ $=$ $t$ $+$ 1 \;
 }
 \caption{Triplet Similarity Embedding: $d_{in}=512$, $d_{out}=128$}
 \label{algo:1}
\end{algorithm}

\begin{figure}
\includegraphics[width=0.5\textwidth]{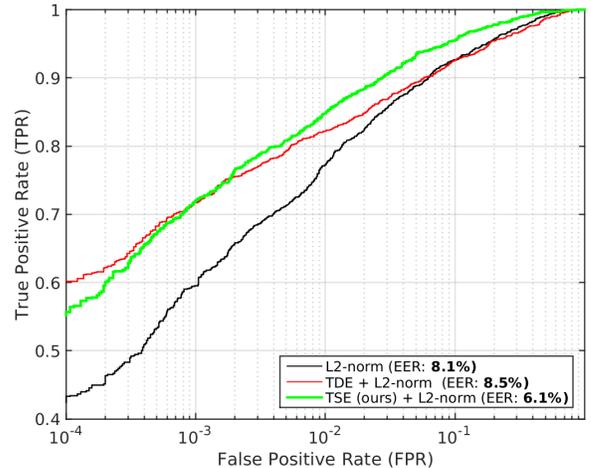}
\caption{Performance improvement on IJB-A split 1}
\label{fig:tl}
\end{figure}

The technique closest to the one presented in this section, which is used in recent works (\cite{facenet15},\cite{parkhi15}) computes the embedding \mbf{W} based on satisfying a distance constraint:
\begin{align*}
\underset{\mbf{W}}{\text{argmin}} \sum_{ {a,p,n} \in \mathbb{T}} max\{0,\alpha + (a-p)^T\mbf{W^TW}(a-p)- \\
				(a-n)^T\mbf{W^TW}(a-n) \}
\tageq \label{eq:dist}
\end{align*}
To be consistent with the terminology used in this paper, we call it Triplet Distance Embedding (TDE). To compare the relative performances of the raw features before projection, TDE and TSE (proposed method), we plot the traditional ROC curve (TAR (vs) FAR) for split 1 of the IJB-A protocol for the three methods in Figure \ref{fig:tl}. The Equal Error Rate (EER) metric, which is a popular measure to compare classification systems is specified for each method. The performance improvement due to the triplet similarity embedding (TSE) is very significant, especially at regions FPR$=\{10^{-4},10^{-3}\}$.

Although the TDE method produces slightly better performance at lower FAR's, the EER value for TDE is worse than using the raw features. On the other hand, the proposed TSE formulation reduces the EER by \textbf{19.7}\% which is a significant improvement. We observed a similar behaviour for all the ten splits of the IJB-A dataset.

%% file: results3.tex
\section{Experimental setup and Results}\label{sec:results}

In this section we evaluate the proposed method on the recently introduced unconstrained face verification dataset IARPA Janus Benchmark-A (IJB-A) \cite{ijba15}. We compare our results with state of the art methods and show that the proposed method always outperforms them while requiring less time to train.
\subsection{IJB-A dataset}
\begin{figure}
\includegraphics[width=0.5\textwidth]{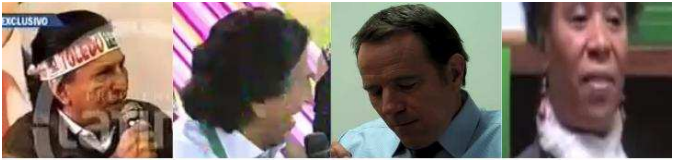}
\caption{Images from the IJB-A dataset}
\label{fig:ijba-sample}
\end{figure}

The IARPA Janus Benchmark-A (IJB-A) contains 500 subjects with a total of 25, 813 images (5,399 still
images and 20,414 video frames). The faces in the IJB-A dataset contain extreme poses and illuminations, much harder than LFW \cite{lfw}. An additional challenge of the IJB-A verification protocol is that the template comparisons include image to image, image to set and set to set comparisons. In this work, if a given test template consists of more than one image, we flatten the template by taking an average of the image features to produce one feature vector. Some sample images from the IJB-A dataset are shown in Figure \ref{fig:ijba-sample}.
\subsection{Pre-processing}
In the training phase, given an input image, 68 fiducial points are extracted using the pipeline described in \cite{iccv15}, out of which six keypoints are used for alignment. In the test phase, we use the three keypoints (left eye, right eye and nose) provided to us through the metadata provided with the IJB-A data, in order to facilitate comparisons with related methods (\cite{wang15},\cite{iccv15}). We use the obtained keypoints to align the image to a canonical view using  a similarity transform. In the IJBA dataset, there are several faces in profile views for which all three keypoints cannot be identified. For these cases, which is almost half of the test data, we crop the faces using the given bounding box information and process them without alignment. Our pre-processing procedure is kept simple to highlight the performance improvement provided by the proposed deep architecture and the triplet similarity embedding.
\subsection{Parameters and training times}
The training of the proposed deep architecture is done using SGD with momentum, which is set to 0.9 and the learning rate is set to 1e-3 and decreased uniformly by a factor of 10 every 40K iterations. The weight decay is set to 5e-4 for all layers. The training batch size is set to 150. We perform finetuning for each split using the training data provided for each split. The finetuning procedure takes 2 hours per split. For the triplet optimization, the alpha value is set to 0.1, which is decided based on a validation set of the first split of IJB-A. The training time for our deep network is 20 hours and the time to obtain the triplet embedding is 10-15 mins per split. During deployment, the average enrolment time per image including pre-processing, alignment and feature extraction is 8ms. The experiments are conducted using an NVIDIA TitanX GPU. 

%
%

\subsection*{IJB-A-Results}
\begin{Table}
\resizebox{1.0\textwidth}{!}{
\begin{tabular}{|c||c||c||c||c|}
 \hline
 \multicolumn{5}{|c|}{IJB-A: Verification Results (TARs at FARs)} \\
 \hline
 FAR & GOTS \cite{ijba15} & \cite{wang15} & \cite{iccv15} & Proposed \\
 \hline
  1e-4 & - & -  & - &\textbf{0.41 $\pm$ 0.08}\\
  1e-3 & 0.20 $\pm$ 0.008  & 0.51 $\pm$ 0.06  & - &\textbf{0.59 $\pm$ 0.05}\\
 1e-2 & 0.41 $\pm$ 0.014  & 0.73 $\pm$ 0.03  & 0.79 $\pm$ 0.04 &\textbf{0.79 $\pm$ 0.03}\\
 1e-1 & 0.63 $\pm$ 0.012  & 0.89 $\pm$ 0.01 & 0.95 $\pm$ 0.01 &\textbf{0.945$\pm$ 0.002}\\
 \hline
\end{tabular}
}
\\ 
\resizebox{1.0\textwidth}{!}{
\begin{tabular}{|c||c||c||c||c|  }
 \hline
 \multicolumn{5}{|c|}{IJB-A: Identification Results} \\
 \hline
Acc. & GOTS \cite{ijba15} & \cite{wang15} & \cite{iccv15} & Proposed \\
 \hline
 R1 & 0.44 $\pm$ 0.02  & 0.82 $\pm$ 0.02  & 0.85 $\pm$ 0.02  &\textbf{0.88 $\pm$ .01}\\
 R5 & 0.60 $\pm$ 0.02  & 0.93 $\pm$ 0.01 & 0.94 $\pm$ 0.01 &\textbf{0.95 $\pm$ .007}\\
 \hline
\end{tabular}
}
\captionof{table}{Identification and Verification results on IJB-A dataset. The results are average over 10 splits and the $\pm$ indicates the standard deviation. $'-'$ implies that the result is not reported for that method.}
\label{results}
\end{Table}

\subsection{Discussion}
Table \ref{results} shows the results for the proposed methods compared to existing results for the IJB-A dataset. As can be observed, the proposed method performs comparable in the verification and better in identification metrics. The first column lists the results of the Government off-the-shelf (GOTS) baseline \cite{ijba15} for the IJB-A protocol. The approach in \cite{wang15} uses an ensemble of six models whereas \cite{iccv15} uses a model trained from scratch using the CASIA-WebFace dataset over 6-7 days. It should be noted that the current work uses a TitanX GPU which is twice as fast as the standard K40 GPU which is used by the other methods. Despite accounting for this difference, we still realize a faster training method compared to other approaches due to our training strategy that involves initializing part of our network with pre-trained weights, which are publicly available. Since the proposed \textbf{TSE} approach maps the data to a more discriminative low-dimensional space, it guarantees better performance with good potential for post processing operations like hashing and visualization. It should be noted that similar to \cite{wang15} and \cite{iccv15}, we perform our experiments in a closed set scenario i.e. we ignore the probe subjects which do not appear in the gallery. The split-wise results of our system and changes in performance at each stage of deployment are discussed in the supplementary material.

%% file: conclusion3.tex
\section{Conclusion and Future Work}\label{conclusion}
In this paper, we proposed a deep CNN-based approach coupled with a low-dimensional discriminative embedding learnt using triplet similarity constraints in a large margin fashion. The proposed pipeline enables a faster training time and improves face verification performance especially at low FAR's. We demonstrated the effectiveness of the proposed method on the challenging IJB-A dataset and achieved state of the art performance. For future work, we plan to embed our TSE approach into training the deep network.